# Mathematical Morphology in Machine Learning

Érick Oliveira Rodrigues[a] and Aura Conci[b]
*Department of Computer Science*
*[a]Universidade Federal de Itajuba, [b]Universidade Federal Fluminense*
*Niteroi, Rio de Janeiro - Brazil*
*Email: erickr@id.uff.br, aconci@ic.uff.br*

***Abstract*—This work introduces concepts from mathematical morphology, an established theory in visual computing, to the field of machine learning. Mathematical morphology operations are sensitive to shapes and density, aspects that are not sufficiently exploited in usual machine learning techniques. A clusterization algorithm that is based on morphological reconstruction is proposed. In this case, the algorithm is faster than state-of-the-art techniques while providing clusterization results that preserve shapes and cluster density. The proposed clusterization scheme possesses a number of unique features such as an intrinsic sense of maximal clusters that can be created, provides a means of removing noise from datasets without extra processing costs, enables a vast amount of growth patterns that are controlled by structuring elements, etc. Furthermore, a novel distance metric that combines Minkowski and Chebyshev distances and is very efficient in terms of morphological dilations is proposed. This combined distance not only achieves very efficient times in neighbourhood iteration tasks in $\mathbb{Z}^2$, but also achieves good results when coupled with the k-Nearest Neighbours (k-NN) classifier. In fact, the proposal was approximately 1.3 times faster than Manhattan and 329.5 times faster than Euclidean in discrete neighbourhood iterations. The distance was evaluated using the k-NN classifier, testing its performance with a total of 33 datasets from the UCI repository and comparing its efficiency to 14 other distances. The proposed distance obtained accuracies that were better than the average more often than its counterparts (in 26 cases out of 33), and also obtained the best accuracy more frequently (in 9 out of 33 cases). Furthermore, morphological classifiers are proposed. No classifier in the current literature is able to properly model shape, density and fractal information in datasets apart from this novel proposal.**

## 1. Introduction

Mathematical Morphology (MM) uses set theory to analyse and process graphical objects. Some of its mostly famous operations alter their size, shape and convexity. MM operations are commonly defined as set operations. A dilation, for instance, uses a structuring element (also a set) to displace another set by summing up each pair of elements of both sets and unites the outcome to the current configuration.

Mathematical morphology is commonplace in the field of visual computing. Unfortunately, this is not the case for machine learning. Some works use a fairly limited amount of MM operations but are not entirely based on mathematical morphology, they use heuristics when it comes to connecting instances to their associated cluster. To the best of our knowledge, no work has ever proposed any classification technique that relies nor uses mathematical morphology techniques in classification. Thus, classification and clusterization methods that adhere to mathematical morphology are proposed and analysed herein.

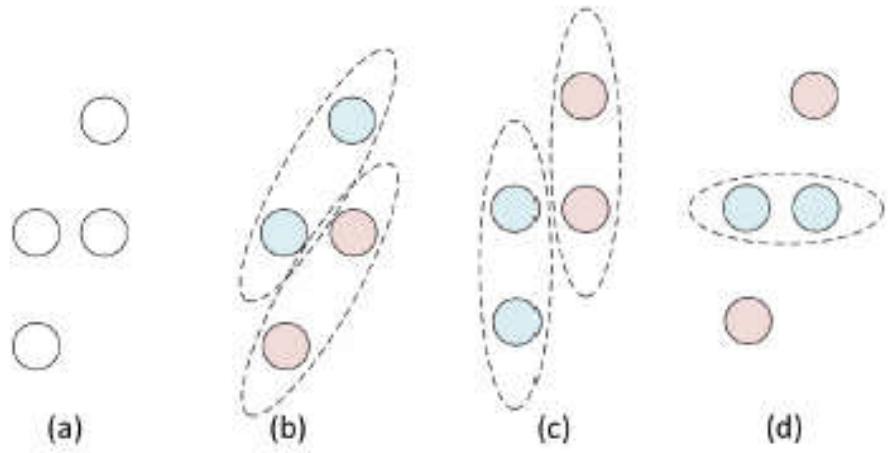

Figure 1: Clusterization results for an arbitrary configuration.

Besides coupling mathematical morphology and machine learning, we developed a new distance metric that is efficient in terms of computing dilations in 2D discrete spaces that can be used in low-level environments such as GPUs. This distance is a metric as long as some restrictions to its parameters are preserved.

Let us suppose that we want to separate the input data shown in Figure 1-(a) in two clusters. Most clusterization algorithms would either produce the results shown in (b) or (c). Although this reasoning makes sense in terms of spatial disposition, it disregards density. The two points in the middle are closely related, i.e., they are dense. In density-based clusterizations these aspects are exploited in a particular fashion.

A possible density-based solution is the one shown in Figure 1-(d), where, in this case, the two data points at the center belong to the same cluster and the remaining is placed in a complement cluster or is treated as noise. The clusterization method proposed in this work uses morphological reconstruction to produce an actual density-based clusterization similar to this.

The proposed algorithm ended up being more efficient than state-of-the-art methods while also being very unique.

This paper is associated to the PhD thesis of Érick Oliveira Rodrigues, former PhD student at Universidade Federal Fluminense and current Adjunct professor at Universidade Federal de Itajuba. Advisor: Aura Conci.

The following list of features inherent to the proposed clustering technique is as follows:

1) Sense of maximal clusters that can be created (regardless of the $k$ parameter), where the $k$ parameter fulfils the same role as $k$-Means, i.e., limiting the amount of clusters.
2) Computationally efficient algorithm as it is amenable to parallelization.
3) Uses structuring elements that alter the way clusters are formed and grouped.
4) Provides a means for removing noise without introducing extra processing costs.

A possible approach for constructing classifiers that use dilations is starting with a set of training instances whose labels are known and growing them, filling the space around these instances in order to cover a wider range of unlabelled instances. As these instances grow, intersections will occur and either the growth stops or a voting scheme should be responsible for choosing the correct label when intersections occur.

Implementing these classifiers brought us to another contribution. We started coding our algorithms in Graphics Processing Units (GPU) to exploit the parallel potential of morphological techniques. That is, assuming the instance space is a $p$-dimensional grid, positions of the grid near positions that contain instances are iteratively marked as the model grows. Each grid cell can be grown in parallel, and therefore, as the GPU paradigm relies on processing the same instruction in parallel with different data, it turns out to be suited for growing several grid positions at once, reducing processing times.

We head towards another issue. The neighbourhood of the grid should be iterated respecting a set of rules. However, how is it possible to obtain an Euclidean-like iteration process without using kernels in the GPU? That is, how is it possible to iterate through the neighbourhood in a discrete environment respecting the Euclidean distance? Kernels are slow. Furthermore, kernels of indefinite sizes are required in this case. We must also acknowledge the fact that grids are discrete spaces. This issue brought us to another solution, which is more similar to the Euclidean distance in terms of information while being outstandingly faster to compute when compared to Manhattan and Euclidean distances.

In 2 dimensions, the proposed distance metric resembles an octagon, while Chebyshev and Manhattan resemble a square. An octagon is closer to the circle obtained with the Euclidean distance. The 2D discrete solution for this distance was developed first. After months of calculations, the distance was successfully associated to a combination of Chebyshev and Minkowski distances, when generalized to $p$-dimensions. An extensive search throughout the literature was performed, which indicates that no work has ever proposed the combination of these distances, and most importantly, no work has ever associated the proposed iterative algorithm (that iterates 2D neighbourhoods) to this combined distance.

In summary, we provide a machine learning framework based on mathematical morphology that is capable of dealing with data that contains fractal information, shapes and density. In addition, we also propose a novel distance metric that is very efficient in terms of morphological dilations. We provide classification experiments and measure the efficiency of the distance in discrete neighbourhood iterations. Two works that originated from this proposal can be found in [1], [2].

## 2. Literature Review

Postaire et al. [3] proposed the first approach to mathematical morphology clusterization. Although not defined by the authors in their original work, these types of algorithms were later termed Binary Morphology Clustering Algorithms (BMCAs) [4], [5]. Figure 2 illustrates how the process of finding cluster "cores" work.

Opening and closing operations are successively applied to data until reaching connected core clusters, as shown in Figure 2-(d). Postaire et al. [3] defines the cluster as a well-connected subset in the data space, which can virtually be of any shape and size. Once the cluster cores are generated, they are considered to be prototypes. The remaining data points are assigned to their respective cluster by means of the nearest neighbour classification rule.

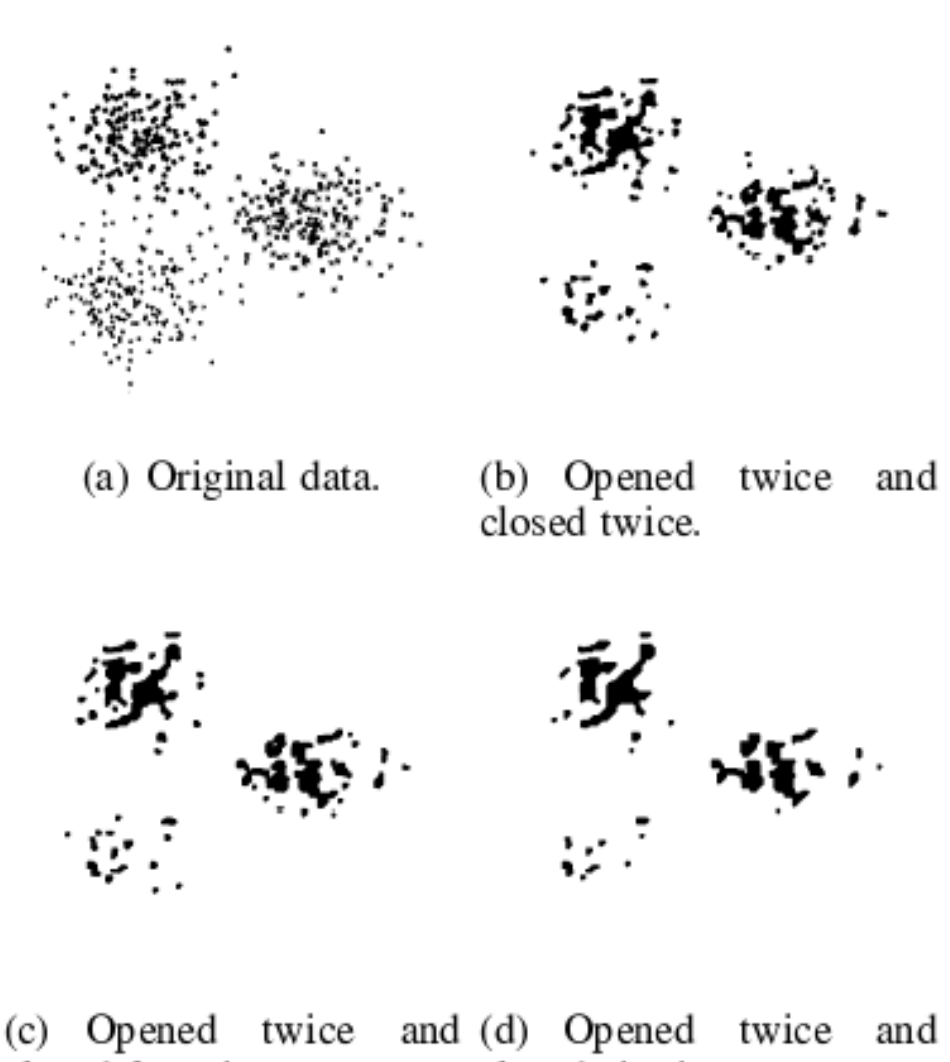

(a) Original data. (b) Opened twice and closed twice.

(c) Opened twice and closed four times. (d) Opened twice and closed six times.

Figure 2: Cluster core extraction using BMCAs.

BMCAs [3], [4], [5], [6] rely entirely on binary operations, as opposed to the algorithm proposed in this work, which works with grey level dilations/image reconstruction. A substantial and fundamental difference is that we do not compute nor locate cluster cores. In our case, elements are promptly assigned to a cluster index as the processing advances. This introduces a noise-classification pattern with no extra processing burden, as noise is segmented in small independent clusters. Furthermore, the clusterization results are very different while also providing a means of limiting the amount of clusters.

## 3. Distance Proposal

The Euclidean distance requires a quantization/approximation process to be computed in discrete environments. The distances can be pre-computed and stored in a kernel or processed in real time. However, storing large amounts of data in kernels consume a significant amount of memory when computing large distances. Accessing this data also impacts negatively on the overall processing performance. In the GPU, for instance, these kernels would have to be located in global memory, whose access is slow. Computing kernels in real time is also very inefficient. Although Manhattan and Chebyshev are faster to compute, they are "square" distances.

On average, the Chebyshev distance iterates neighbourhoods up to distance $D = 2500$ in 0.098 seconds. $D$ represents the size of the discrete neighbourhood around the pixel that is being iterated. The Manhattan distance, on the other hand, takes approximately 0.146 seconds to iterate through the same neighbourhood. The Euclidean distance, at last, requires a total of 36.125 seconds, which is significantly worse.

When it comes to optimization, especially in GPU computing, these differences on the performance impact on the overall processing times even more. Besides, the Manhattan distance does not add any information to the neighbourhood iteration when compared to the Chebyshev distance. The Manhattan distance is similar to the Chebyshev distance yet rotated.

On the other hand, using the Euclidean distance implies a huge computational burden, where its implementation is approximately 368 times slower than Chebyshev for $D = 2500$, and it gets even worse as $D$ increases. This work proposes an intermediate distance metric between Chebyshev and Euclidean that adds information, as opposed to Manhattan, and is far more efficient when it comes to neighbourhood iterations.

This distance (its mathematical formulation is shown in Equation 1) was established while attempting to improve the neighbourhood iterations of morphological classifiers in GPUs, which heavily rely on low level operations. As previously stated, a metric closer to the Euclidean distance was desired, but the reported time burden was incompatible with practical cases.

This proposed iteration achieves Chebyshev-like run times and obtains an iteration fashion that is closer to the Euclidean. The visual result of the proposed distance in $\mathbb{R}^2$ for $p = 1$ is an octagon (Chebyshev and Manhattan distances produce squares). That is, the octagon is much closer to the circular pattern produced by the Euclidean distance than the squares produced by the Manhattan and Chebyshev distances. Table 1 shows the processing times obtained with each one of the algorithms as the size of the neighbourhood $D$ is increased. These results were obtained using Java 7 and an Intel i7-7700HQ, clocked at 2.8 GHz averaged over 100 runs.

TABLE 1: Processing times (s) for each distance.

| Sizes (D) | Chebyshev | Manhattan | Euclidean | **Rodrigues** |
|---|---|---|---|---|
| 500 | 0.0022 | 0.0024 | 0.3069 | 0.0022 |
| 1000 | 0.0109 | 0.0116 | 2.3290 | 0.0115 |
| 1500 | 0.0295 | 0.0407 | 7.8285 | 0.0318 |
| 2000 | 0.0583 | 0.0829 | 18.5112 | 0.0624 |
| 2500 | 0.0985 | 0.1460 | 36.1253 | 0.1050 |
| 3000 | 0.1517 | 0.2228 | 62.2829 | 0.1599 |
| 3500 | 0.2377 | 0.3397 | 98.8083 | 0.2454 |
| 4000 | 0.3031 | 0.4672 | 147.6250 | 0.3166 |
| 4500 | 0.4030 | 0.5923 | 204.2881 | 0.4243 |

Table 1 demonstrates that the proposed distance is in fact faster than the Manhattan and Euclidean distances. On average, it was 1.3 times faster than Manhattan and 329.5 times faster than Euclidean. In what follows, a formal mathematical definition of this distance is presented.

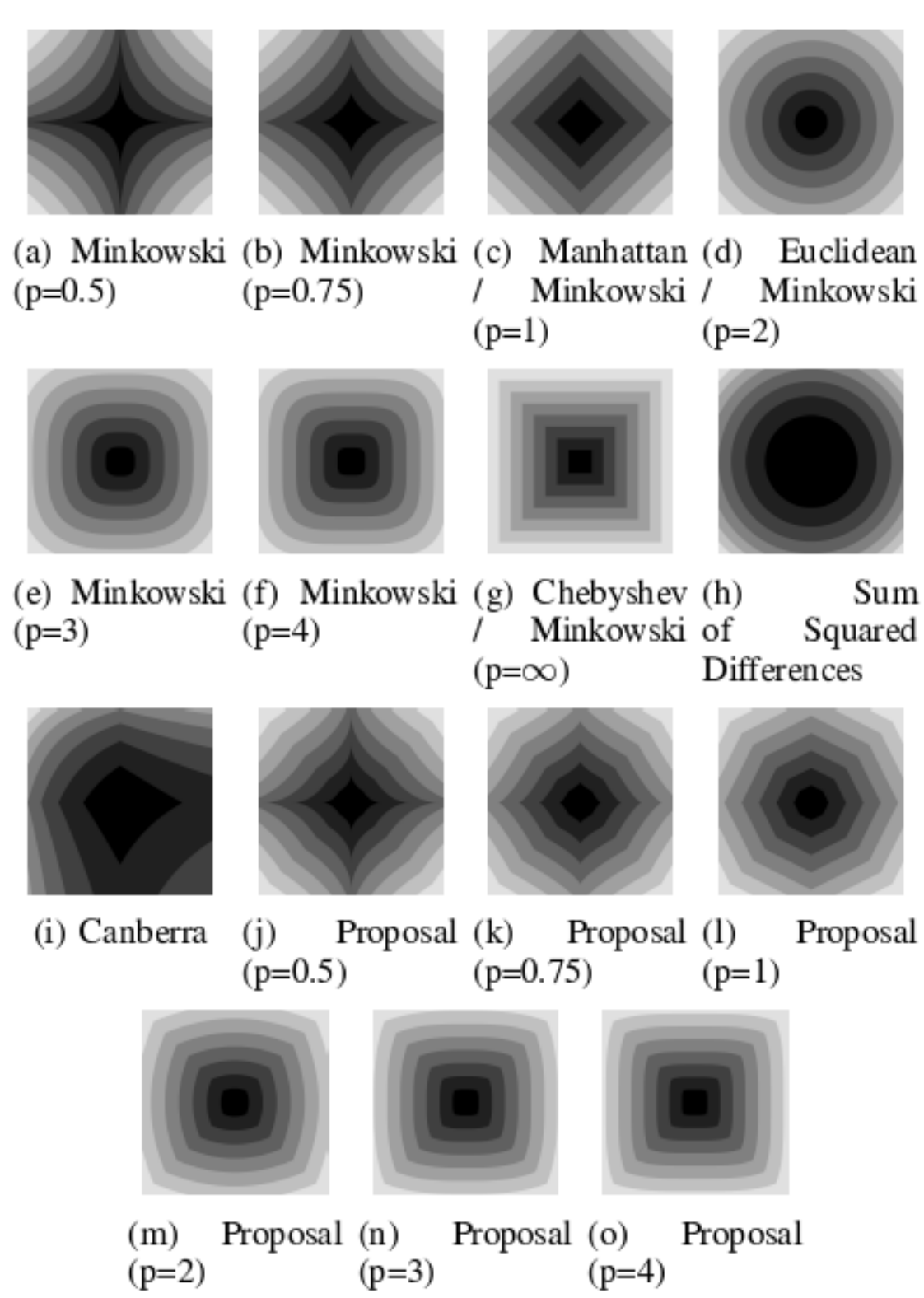

Figure 3: Distances in $\mathbb{R}^2$. The distances are computed from the central element in the image. Lighter shades of gray indicate greater distances.

**3.0.1. Definition.** The proposed distance is a combination of Chebyshev and Minkowski distances, weighted by $w_1$ and $w_2$, as shown in Equation 1. As $w_1$ increases in regard to $w_2$, the distance becomes more like Minkowski. On the contrary, when $w_2$ increases it converges towards Chebyshev.

**Definition** The proposed distance is defined as:

$$d_{w_1,w_2,p}(x,y) = w_1 \sqrt[p]{\sum_{i=1}^{n} |x_i - y_i|^p + w_2 \max_{i=1}^{n} |x_i - y_i|} \tag{1}$$

As two metrics are being summed up, it is straightforward to infer from the formulation that as long as $w_1, w_2 > 0$ and $p \geq 1$, all the metric conditions are satisfied, including the triangle inequality, proved in our thesis.

**3.0.2. Graphical Analysis in $\mathbb{R}^2$.** Figure 3 shows how the proposed distance compares to other distances in $\mathbb{R}^2$. The proposed distance is in fact a mixture of Chebyshev and Manhattan (for p = 1), as shown in Figure 3-(l). It is possible to argue that the distance is an intermediate step between Manhattan/Chebyshev and Euclidean, and therefore it adds information in relation to Manhattan and Chebyshev. However, it requires far less processing power in regard to Euclidean or even Manhattan. For $p < 1$, the respective distances are not metrics. The condition $w_1 = w_2 = 1$ is respected for all the distances in this figure.

## 4. Morphological Clustering Proposal

We will now shift our attention to the clustering proposal. This section focuses on describing a single clusterization algorithm, called k-Morphological Sets (k-MS), which is capable of segmenting and grouping data respecting shapes and cluster density [2]. The method is heavily based on morphological reconstruction.

Before heading to how the algorithm works, we define how data is preprocessed. At first, an image or grid $G$ is constructed. $G(x,y)$ returns the index value $x \times y$ in case there is an instance at position $(x,y)$ or 0 otherwise. An image mask $M(x,y)$ returns 1 in case an instance is positioned at $(x,y)$ and 0 otherwise.

Once image $G$ is created, a reconstruction $R_M^B(G)$ is applied. A structuring element $B$ whose element values are 0 (including the central element marked with $\times$ is used along with size 1 for the $sup$ metric. The size of the structuring element can be altered, which leads to distinct types of clusterizations. The shape of the element can be modified as long as it contains all directions (e.g., left, right, up and down for the case of 2 dimensions). If this is not the case, the algorithm may reach a potential deadlock.

A structuring element that contains only zeros implies that the dilation would just spread the biggest values through some parts of $G$ (limited by mask $M$), biased by the size and shape of the structuring element. Therefore, after the reconstruction, a clusterized image is obtained.

This morphological reconstruction does not separate input data in a prescribed number of clusters. Instead, the algorithm separates the data based on “connectivity”, relying on the size and shape of the structuring element. Therefore, we introduce an internal variable $\delta$ that is responsible for increasing the size of the structuring element $B$. That is, in the beginning of the clusterization, the size of $\delta$ is minimal, as depicted in Figure 4-(a). Eventually, the algorithm reaches an idempotent configuration. Therefore, we gradually increase the size of the structuring element until it reaches the remaining grey instances at the bottom of the image, shown in (b). At each increment of $\delta$, the algorithm checks for the condition of idempotence, and if it is still idempotent, it increments $\delta$. Otherwise, the algorithm resets $\delta$ to the minimum, such as shown in (c). At that stage, the grey level value of the instances at the bottom reaches the upper instances, passing the index of the bottom cluster to them (d). Figure 4 depicts a hypothetical situation where $k = 1$, i.e., just one cluster is desired.

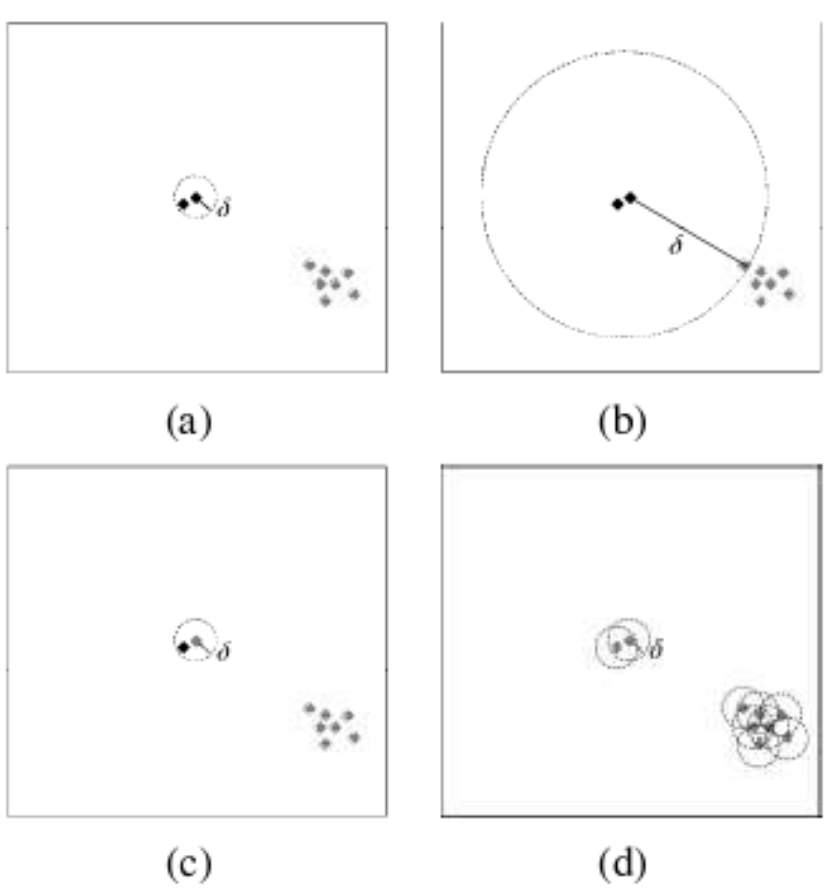

Figure 4: Some steps of the proposed k-MS algorithm.

When it comes to the proposed algorithm, $\delta$ is a factor that multiplies the size of the used structuring element $B$. In the beginning of the algorithm, $\delta$ is set to 1. A morphological reconstruction operation is then performed on image $G$ using the structuring element $B$ and mask $M$.

Later, the number of unique values in $G$ is calculated, which corresponds to the number of clusters created in $G$ at that moment. This computation is performed using early breaks, as shown in Algorithm 1, to speed up the processing. If $G$ has more than $k$ clusters, the algorithm continues to the next step, where $\delta$ is incremented. $G$ is dilated once. If $G$ is idempotent, then $\delta$ is incremented until $G$ is not idempotent. The idempotence is overcome when an index that belongs to a certain cluster reaches another with a different index. When this happens, the algorithm resets $\delta$ to 1 and starts the same process all over again. With less than $k$ or exactly $k$ clusters in $G$, the algorithm terminates.

This morphological application also extends to classification problem, more details can be found at this paper of ours [7].

### 4.1. Experiments

This section compares the visual results obtained in a noisy publicly available dataset [8]. In this case, we compare k-MS to clusterization algorithms that are sensitive to density and shapes. The behaviour of k-MS is

**Algorithm 1:** Low-level k-Morphological Sets algorithm.

**Data:** $B$ is the structuring element, $k$ is the desired maximum number of clusters to be created and $R$ the reconstruction function

1 **begin**
2 finished $\leftarrow$ false;
3 $\delta \leftarrow 1$;
4 **while** *!finished* **do**
5 idempotent $\leftarrow$ true; finished $\leftarrow$ true;
6 Reset or remove values from the kArray (the array contains at most $k$ different values);
7 **for** *every instance or data point $p$ in the input dataset* **do**
8 $p_{aux} \leftarrow p$;
9 If the surroundings of pixel $p$, respecting the structuring element $B$ and $\delta$, contains a higher value than $p$, then $p$ receives the highest surrounding value;
10 **if** *!idempotent* **then continue**; `// first early break`
11 **if** $p_{aux} \neq p$ **then** idempotent $\leftarrow$ false;
12 **if** *!finished* **then continue**; `// second early break`
13 **if** *kArray does not contain the value of $p$* **then**
14 **if** *kArray is fully populated* **then** finished $\leftarrow$ false;
15 **else** Add the value of $p$ to kArray at a vacant position;
16 **if** *idempotent* **then** $\delta \leftarrow \delta + 1$;
17 **else** $\delta \leftarrow 1$;
18 finished $\leftarrow$ finished **and** idempotent;
19 Returns all the instances $p$ where each one is indexed with a unique cluster index (amount of clusters $\leq$ k);

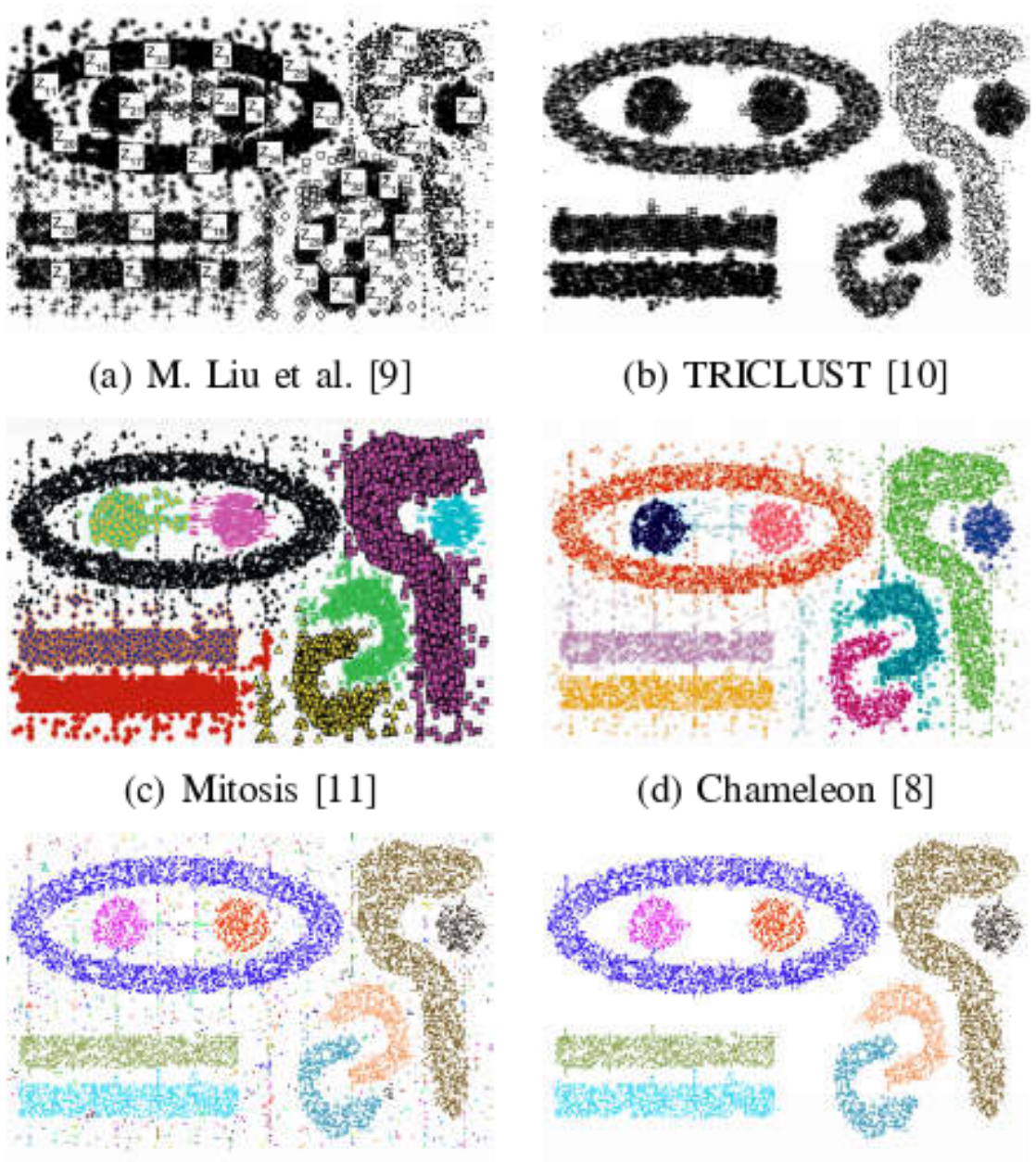

(a) M. Liu et al. [9] (b) TRICLUST [10]

(c) Mitosis [11] (d) Chameleon [8]

(e) k-MS - Proposal ($k = 450$) (f) k-MS-Proposal (Rem. Noise)

Figure 5: Comparison of the visual results of k-MS and other clustering algorithms in finding genuine morphological clusters.

also analysed using different input parameters. Finally, in the third group of experiments, we perform an extensive time analysis, comparing the run times obtained by the different implementations of k-MS with other clusterization algorithms.

The clusterizations performed with k-MS, Chameleon, Mitosis, TRICLUST and M. Liu et al. algorithms on the dataset provided by Karypis et al. [8] are analysed and compared. Figure 5 shows the clusterizations obtained with each algorithm.

The clusterizations of Mitosis and M. Liu et al. recognize the exact amount of genuine clusters (total of 9 clusters), such that all noise in the image is clusterized together with these genuine clusters. This outcome can be bad in some occasions such as when noise is expected to be removed or when it should not be aggregated in a genuine cluster. A practical example of this is image data that was collected by physical sensors.

On the other hand, the clusterization generated by Chameleon recognizes 9 genuine clusters as well as 3 other clusters that contain outliers. Chameleon and k-MS are the only two algorithms capable of segregating outliers in different clusters. k-MS goes even further, it indicates whether the clusters can be formed given a predefined $k$ amount of clusters.

The result shown in Figure 5-(e) shows the clusters obtained with k-MS algorithm, where each cluster is depicted in a different color. In noisy datasets, the value of $k$ should be large, otherwise k-MS would not be able to recognize the genuine clusters in the dataset. Furthermore, if the dataset is too sparse, it is also recommended to perform dilations on the dataset before applying k-MS, since the clusterization would be more accurate and converge faster.

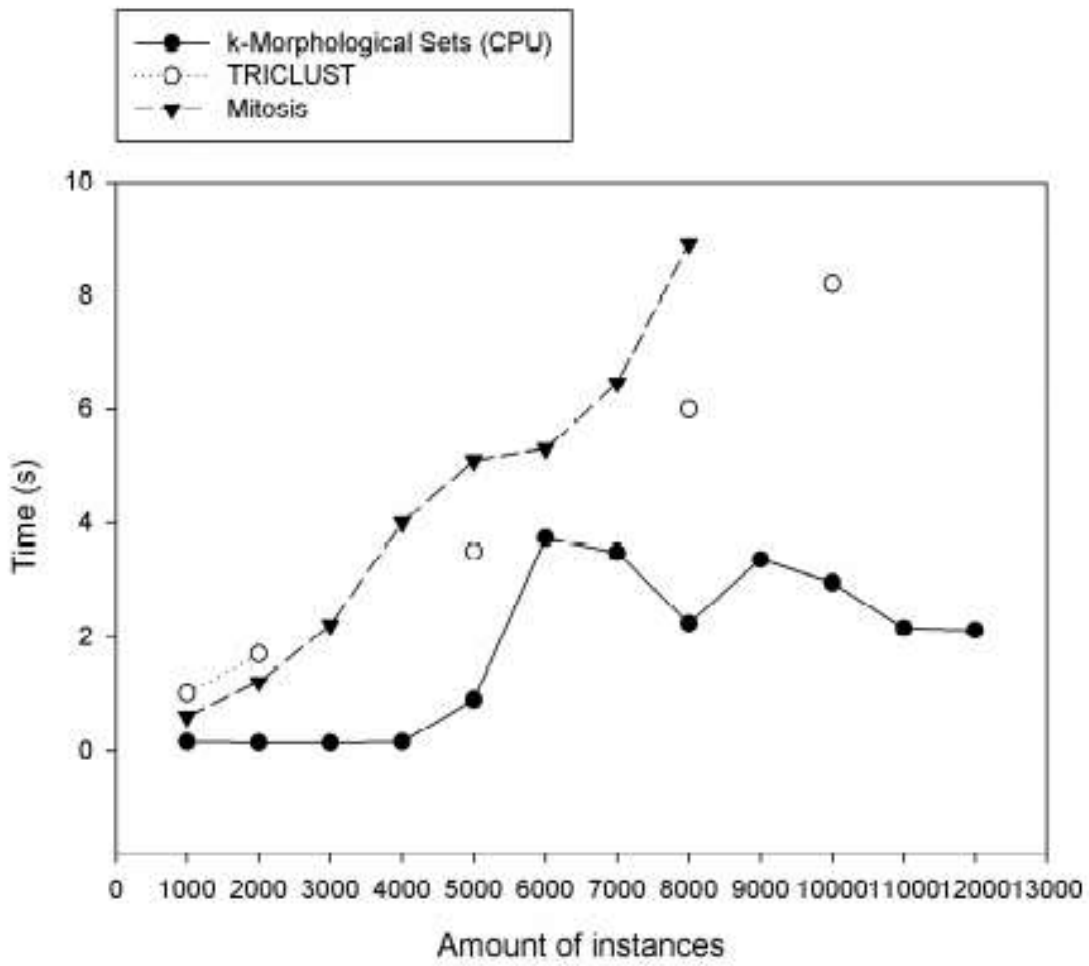

Figure 6: Run times (s) comparison for varying numbers of instances.

**4.1.1. Run Time Analysis.** An extensive time analysis experiment is performed in this section. In order to compare and measure the time performance on datasets with a large amount of instances, we considered the GPU-oriented k-Means algorithm presented in [12], which is based on [13] and [14]. However, this implementation could not handle the size of the instances we evaluated. Therefore, our comparisons on this section address just the implementations of [13], which is a parallel k-Means for the CPU using

OpenMP. We tested the sequential k-Means provided by them as well but it was always slower than the parallel one, and therefore it was disregarded.

k-Means starts faster for low values of $k$ but is overcome in every occasion after a particular size of $k$. As the size of $k$ increases, so does the run time of clusterization algorithms in general. However, due to the early break in k-MS, the run times obtained with k-MS may even decrease, regarding the CPU, as $k$ increases. This happens because the clusterization converges faster with higher values of $k$ (less morphological reconstructions), and the early breaks reduce the time of the clusters verification. Furthermore, the GPU implementation is only viable if the amount of instances is very large and $k$ is small. Small values of $k$ reduce the amount of atomic operations that have to be performed, which reduces the processing time of the algorithm.

## 5. Conclusion

k-MS outperformed all state-of-the-art clustering algorithms that are sensitive to cluster shape and density in terms of processing times. Besides, it is the only algorithm that has an intrinsic sense of maximal clusters that can be created. Given a predefined $k$ variable, the algorithm outputs either $k$ or less clusters, depending on the data configuration. When density is regarded, it is not possible to separate high density clusters in some occasions. Let us suppose connected data points forming a shape of a circle. In this case, if $k > 1$, the algorithm would still segregate the data in a unique cluster. In addition, it is also the only algorithm in the literature that provides a noise removal paradigm without requiring extra processing costs.

An extra major contribution consists of a metric that was coined on the process of iterating image neighbourhoods using GPUs (in order to perform image dilations). GPUs rely on low level instruction environments and therefore complex data structures are not practical. This proposed distance is closer to Euclidean in terms of neighbourhood iteration, when compared to Chebyshev and Manhattan, while being faster to compute. In fact, when plotted in $\mathbb{R}^2$, the proposed metric displays an octagon, which is closer to the circle displayed by the Euclidean metric. As the proposal consists of a sum of two metrics, we prove that it is a metric as well [1].

An extensive experiment regarding 33 datasets of the UCI repository was conducted in order to evaluate the efficiency of the proposed distance using k-NN. This novel distance achieved accuracies better than the average more frequently than its counterparts (26 cases in 33). Furthermore, it obtained the best accuracies among the evaluated distances more often (9 cases in 33) [1].

## 6. List of Publications

This is a list of publications that originated or are related to this PhD thesis, which includes high reputable journal publications such as Pattern Recognition and Pattern Recognition Letters: [1], [2], [7], [15], [16], [17], [18], [19], [20], [21], [22], [23].

## References

[1] E. O. Rodrigues, "Combining minkowski and cheyshev: New distance proposal and survey of distance metrics using k-nearest neighbours classifier," *Pattern Recognition Letters*, vol. 110, 2018.

[2] E. O. Rodrigues, L. Torok, P. Liatsis, J. Viterbo, and A. Conci, "k-ms: A novel clustering algorithm based on morphological reconstruction," *Pattern Recognition*, vol. 66, pp. 392–403, 2016.

[3] J. G. Postaire, R. D. Zhang, and L. Botte, "Cluster analysis by binary morphology," *IEEE Transactions on Pattern Analysis and Machine Intelligence*, vol. 15, pp. 170–180, 1993.

[4] E. C. Pedrino, M. C. Nicoletti, J. H. Saito, L. M. V. Cura, and V. O. Roda, "A binary morphology-based clustering algorithm directed by genetic algorithm," *IEEE International Conference on Systems, Man, and Cybernetics (SMC)*, 2013.

[5] H. Luo, F. Kong, K. Zhang, and L. He, "A clustering algorithm based on mathematical morphology," *Proceedings of the 6th World Congress on Intelligent Control and Automation*, 2006.

[6] C. S. Chen and C. W. Yeh, "An efficient dilation-based clustering algorithm for automatic optical inspection," *The 11th International Conference on Information Sciences, Signal Processing and their Applications*, 2012.

[7] E. O. Rodrigues, A. Conci, and P. Liatsis, "Morphological classifiers," *Pattern Recognition*, 2018.

[8] G. Karypis, E. H. Han, and V. Kumar, "Chameleon: Hierarchical clustering using dynamic modeling," *Computer*, vol. 32, pp. 68–75, 1999.

[9] M. Liu, X. Jiang, and A. C. Kot, "A multi-prototype clustering algorithm," *Pattern Recognition*, vol. 42, no. 5, pp. 689–698, 2009.

[10] D. Liu, G. V. Nosovskiy, and O. Sourina, "Effective clustering and boundary detection algorithm based on delaunay triangulation," *Pattern Recognition Letters*, vol. 29, pp. 1261–1273, 2008.

[11] N. A. Yousri, M. S. Kamel, and M. A. Ismail, "A distance-relatedness dynamic model for clustering high dimensional data of arbitrary shapes and densities," *Pattern Recognition*, vol. 42, pp. 1193–1209, 2009.

[12] S. Giuroiu, "Cuda k-means clustering," http://serban.org/software/kmeans/.

[13] W. Liao, "Parallel k-means data clustering," available at http://users.eecs.northwestern.edu/wkliao/Kmeans/index.html.

[14] J. MacQueen, "Some methods for classification and analysis of multivariate observations," *Berkeley Symposium on Mathematical Statistics and Probability*, vol. 1, pp. 281–297, 1967.

[15] E. O. Rodrigues, P. Liatsis, L. S. Ochi, and A. Conci, "Fractal triangular search: A metaheuristic for image content search," *IET Image Processing*, 2018.

[16] E. O. Rodrigues, F. Morais, N. Morais, L. Conci, L. Neto, and A. Conci, "A novel approach for the automated segmentation and volume quantification of cardiac fats on computed tomography," *Computer Methods and Programs in Biomedicine*, vol. 123, pp. 109–128, 2016.

[17] E. O. Rodrigues, V. H. A. Pinheiro, P. Liatsis, and A. Conci, "Machine learning in the prediction of cardiac epicardial and mediastinal fat volumes," *Computers in Biology and Medicine*, 2017.

[18] E. O. Rodrigues, L. O. Rodrigues, L. S. N. Oliveira, A. Conci, and P. Liatsis, "Automated recognition of the pericardium contour on processed ct images using genetic algorithms," *Computers in Biology and Medicine*, vol. 87, pp. 38–45, 2017.

[19] E. O. Rodrigues, A. Conci, F. F. C. Morais, and M. G. Perez, "Towards the automated segmentation of epicardial and mediastinal fats: A multi-manufacturer approach using intersubject registration and random forest," *IEEE International Conference on Industrial Technology (ICIT)*, pp. 1779–1785, 2015.

[20] E. O. Rodrigues, F. F. C. Morais, and A. Conci, "On the automated segmentation of epicardial and mediastinal cardiac adipose tissues using classification algorithms," *MEDINFO 2015: EHealth-enabled Health: Proceedings of the 15th World Congress on Health and Biomedical Informatics*, vol. 216, 2015.

[21] E. O. Rodrigues, J. Viterbo, A. Conci, and T. McHenry, "A context-aware middleware for medical image based reports an approach based on image feature extraction and association rules," *IEEE International Conference on Computer Systems and Applications*, 2015.

[22] E. O. Rodrigues, T. M. Porcino, A. Conci, and A. C. Silva, "A simple approach for biometrics: Finger-knuckle prints recognition based on a sobel filter and similarity measures," *International Conference on Systems, Signals and Image Processing (IWSSIP)*, 2016.

[23] E. O. Rodrigues and E. Clua, "A real time lighting technique for procedurally generated 2d isometric game terrains," *Entertainment Computing - ICEC 2015*, vol. 9353, pp. 32–44, 2015.